%% file: AnonymousSubmission2027.tex
\documentclass[letterpaper]{article} 
\usepackage[preprint]{aaai2027} 

\usepackage[hyphens]{url}  
\usepackage{graphicx} 
\usepackage{natbib}  
\usepackage{caption} 
\usepackage{algorithm}
\usepackage{algorithmic}
\usepackage{amssymb}
\usepackage[table]{xcolor} 

\usepackage{newfloat}
\usepackage{booktabs}
\usepackage{amsmath}
\usepackage{tabularx}
\usepackage{tikz}
\usetikzlibrary{arrows.meta,positioning}
\usepackage{multirow}
\usepackage{listings}
\usepackage{adjustbox}
\usepackage[most]{tcolorbox}
\tcbset{
  colback=gray!3,
  colframe=gray!40,
  boxrule=0.4pt,
  arc=2pt,
  left=4pt,right=4pt,top=3pt,bottom=3pt,
  breakable,
  enhanced,
  before skip=0pt,
  after skip=0pt
}

\DeclareCaptionStyle{ruled}{labelfont=normalfont,labelsep=colon,strut=off} 
\floatstyle{ruled}
\newfloat{listing}{tb}{lst}{}
\floatname{listing}{Listing}

\usepackage{booktabs}

\title{\textsc{Latent-IM}: Latent Interaction Management for Speech LLMs}

\author{Adar Avsian, Atahan Dokme, Tony Woo, Larry Heck }
\affiliations{
    Georgia Institute of Technology\\
    \{adar, adokme3, tonywoo, larryheck\}@gatech.edu
}

\begin{document}

\maketitle

\begin{abstract}
Classical spoken dialogue systems often separated dialogue management from response realization: a policy selected the next dialogue action, and a generation component expressed that action. As dialogue systems shift toward LLMs, this decomposition has largely disappeared into the model's hidden representations. We ask whether an LLM-internal analogue of state estimation and action control can be recovered for conversational moves such as acknowledging, checking, querying, explaining, and replying. We formulate move control as two coupled problems: \emph{selection}, predicting the appropriate next move from the dialogue context, and \emph{realization}, causally producing a chosen move at generation time. We introduce \textsc{Latent-IM}, an internal dialogue-management framework that provides a general interface for choosing and deploying conversational moves under different objectives. Here, we use this control to reproduce human move choices, improving average end-to-end move accuracy by 12.5 points over the unsteered backbone while performing comparably to fine-tuning.

\end{abstract}


\section{Introduction}

Classical dialogue systems often modeled conversation as a state--action control problem. A dialogue manager maintained a representation of the conversation state, often a belief state under uncertainty, and a policy selected the next system action. This decomposition was especially influential in Partially Observable Markov Decision Process (POMDP) approaches, where noisy observations were integrated into a belief state and actions were chosen to maximize expected dialogue success \citep{williams2007partially,young2013pomdp}. Modern LLMs instead take an end-to-end approach, generating responses directly from the dialogue history without explicit state management or action policy.  

Although modern LLMs lack explicit state and policy modules, they must still resolve the same underlying control problem: determining what conversational action is appropriate in the current context \citep{shaikh2024grounding,maitra2025dialogue,luo2025clarifymt, avsian2026sneak,nguyen2023generative,veluri2024beyond,lee2025behavior}. This decision must therefore be represented and resolved implicitly within the LLM's hidden activations. This raises our central question: can an internal analogue of state estimation and action control be recovered from these representations and used to govern conversational behavior?

We introduce \textsc{Latent-IM}, a framework for recovering interaction management (IM) from the representations of a frozen speech LLM. We operationalize IM as deciding which local conversational action the model should take and when it should yield the floor. Specifically, we study five moves that support grounding, information exchange, and uncertainty resolution. Mirroring the classical state--action decomposition, \textsc{Latent-IM} separates \emph{selection}, which predicts the next move from residual-stream activations and reaches 0.60 average accuracy, from \emph{realization}, which uses activation steering to express that move and outperforms the strongest baseline by 10.4 percentage points. A complementary turn-boundary direction controls when the response terminates, monotonically varying mean response length from $71.3$ to $10.4$ words across the steering range.


Our contributions are as follows:

\begin{itemize}
    \item We combine activation-based selection and realization into \textsc{Latent-IM}, an end-to-end dialogue-management system that operates without gold moves or backbone updates, outperforming other control baselines and matching supervised fine-tuning. 
    \item We show that a lightweight streaming controller can recover conversational state from frozen speech LLM activations, outperforming transcript-based selectors across three task-oriented dialogue datasets. 
    \item We derive reusable move-specific activation directions that causally realize conversational actions. Directions learned from MapTask transfer without re-estimation to FindTask and CReST and outperform prompting, candidate-selection, and decoding-time control baselines. 
    \item We identify a distinct turn-boundary direction that causally controls response length and integrate it with realization-aware gating to regulate when the model yields the floor. 
\end{itemize}

\section{Related Work}

\paragraph{Dialogue State, Policy, and Conversation Analysis.}
Classical task-oriented dialogue systems framed dialogue management as state tracking followed by action selection. Finite-state and frame-based systems explicitly represented task progress and slot values, while POMDP approaches maintained beliefs over latent dialogue states to support action under uncertainty. Conversation analysis and dialogue-act theory complement this view by describing interaction through sequential actions such as grounding, repair, acknowledgment, clarification, and response \citep{clark1991grounding, traum1992conversation, sacks1974simplest, schegloff1977preference,purver2001means}.  Although classical systems provided an explicit state/action abstraction, they were often tied to domain-specific intents, slots, and templates. Recent work such as NextLat induces belief-like internal states through latent-prediction objectives \citep{teoh2025next}. We instead recover dialogue state from a frozen LLM's residual stream and represent actions as domain-general conversational moves.

\paragraph{LLMs and Activation Steering.}
LLMs can produce fluent dialogue while underproducing grounding and repair, answering prematurely under ambiguity, diverging from human dialogue-act distributions \citep{shaikh2024grounding,maitra2025dialogue,luo2025clarifymt, dokme2026temper}. Prior work has used internal representations to modify model behavior at inference time \citep{li2023inference,todd2024function,panickssery2023steering,avsian2026flex,subramani2022extracting, dokme2026selective, zou2023representation}.  Activation Addition constructs contrastive directions for attributes such as sentiment, topic, and toxicity \citep{turner2024steering}, while methods such as conceptor steering capture richer activation structure \citep{postmus2024steering}. Recent work extends activation steering to speech and audio-language models, including temporal audio attention, speech adaptation, and interruption control \citep{chang2026overcoming,lin2026steering,yegorova2026salsa}. Our work instead targets local, context-dependent conversational actions by separating when a move should be selected from how it is causally realized.

\section{Task \& Data}
\label{sec:task-data}

We study controllable generation of conversational moves in asymmetric,
instruction-oriented task dialogues. Let a dialogue \(D\) be a sequence of turns
\[
D=\bigl((r_1,u_1),\ldots,(r_T,u_T)\bigr),
\]
where \(r_t\) and \(u_t\) denote the speaker role and utterance at turn \(t\). All datasets contain two asymmetric participant roles, which we call \emph{giver} and \emph{follower}. The giver provides task-relevant instructions or information, while the follower acts on that information. We focus on follower turns because they exhibit a richer and more interaction-oriented move inventory. For a follower turn \(t\), the preceding context is
\[
c_t=\bigl((r_1,u_1),\ldots,(r_{t-1},u_{t-1})\bigr),
\]
and the human response \(u_t\) has gold move label
\[
y_t\in\mathcal{M} =
\{\textit{acknowledge},\textit{check},\textit{explain},
\textit{query},\textit{reply}\}.
\]
We map each dataset's annotations into this shared taxonomy, defined in
Table~\ref{tab:moves}. These labels capture domain-general interactional
functions rather than task-specific intents. Turn completion is modeled
separately and is not included in \(\mathcal{M}\).

We decompose move control into two tasks. \emph{Selection} predicts the
appropriate follower move from the dialogue context:
\[
p_\theta(y_t\mid c_t)=s_\theta(c_t),
\qquad
\hat y_t=
\arg\max_{m\in\mathcal{M}}
p_\theta(y_t=m\mid c_t).
\]
\emph{Realization} generates a response conditioned on a target move:
\[
\hat u_t=g_\phi(c_t,y_t^\star).
\]
In oracle-realization experiments, \(y_t^\star=y_t\) is the gold human move. In
the end-to-end setting, \(y_t^\star=\hat y_t\) is selected by the controller.
Thus, selection determines \emph{which} conversational move to make, while
realization determines \emph{how} that move is expressed.

\begin{table}[h]
\centering
\small
\setlength{\tabcolsep}{5pt}
\renewcommand{\arraystretch}{1.08}

\begin{tabularx}{\columnwidth}{@{}lX@{}}
\toprule
\textbf{Move} & \textbf{Definition} \\
\midrule

acknowledge
& Signals that the follower has heard or is ready to continue, without adding
new task information. \\

check
& Verifies the giver's instruction or confirms the follower's current
understanding of the route. \\

explain
& Provides additional task-relevant information, reasoning, or clarification
about the follower's state. \\

query
& Requests information from the giver, including yes/no and wh-questions. \\

reply
& Answers a giver question or check, including yes/no and wh-replies. \\

\bottomrule
\end{tabularx}

\caption{Merged follower-move types.}
\label{tab:moves}
\end{table}

\subsection{Datasets}
We evaluate this formulation on three dialogue datasets. 

\paragraph{MapTask.} Our primary dataset is MapTask \citep{anderson1991hcrc}, a collection of unscripted, task-oriented spoken dialogues in which two participants collaborate to reproduce a route on a map. The giver has a map with a marked route, while the follower has a related but non-identical map and must draw the route from verbal instructions.

\paragraph{FindTask.} We also use FindTask from the RoboHelper corpus \citep{chen-di-eugenio-2013-multimodality}, which contains task-oriented human-robot dialogues in which participants collaborate with a robotic helper to locate objects, exchange task information, and resolve misunderstandings.

\paragraph{CReST.}
Finally, we include the Indiana Cooperative Remote Search Task (CReST) corpus \citep{eberhard-etal-2010-indiana}, in which a director views an indoor map and guides a remotely connected searcher through the environment.

\begin{figure*}[t]
    \centering
    \includegraphics[width=\linewidth]{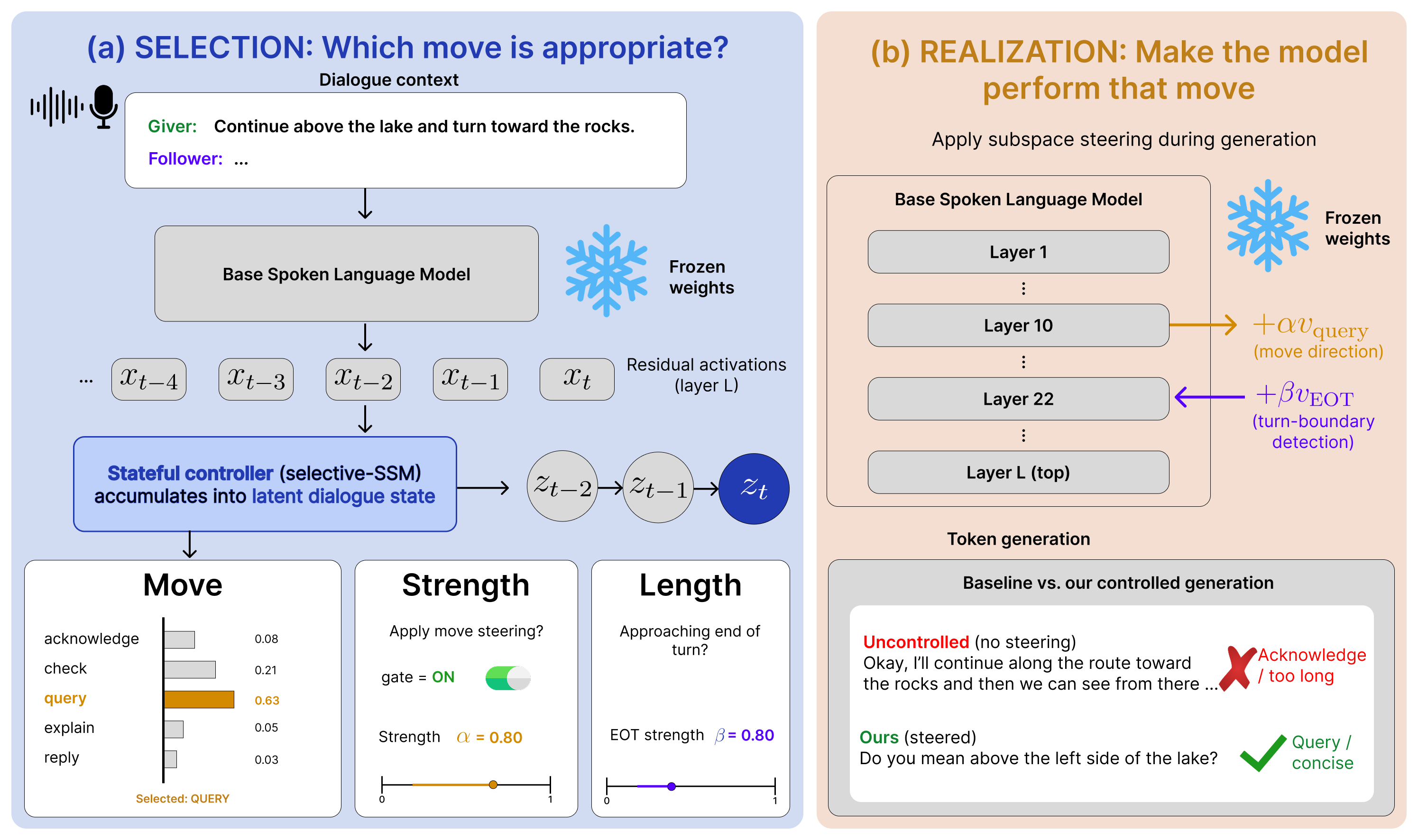}
    \caption{(a) Before generation, the controller predicts the next conversational move, whether and how strongly to apply move steering, and the appropriate turn length (via a turn-boundary strength). These predictions are made without gold labels at inference. In this example, the controller selects QUERY as the next move. (b) During generation, the selected move direction $v_{\textsc{query}}$ and the turn-boundary direction $v_{EOT}$ are injected at selected transformer layers with strengths  $\alpha$ and $\beta$. This causes the frozen backbone to realize the selected move and terminate at an appropriate boundary.}
    \label{fig:main}
\end{figure*}

\section{Method}

\textsc{Latent-IM} recovers a classical dialogue-management pipeline within a frozen speech LLM (Figure~\ref{fig:main}). A recurrent controller reads residual-stream summaries to maintain dialogue state and select the next conversational move, which is then realized through move-specific activation steering rather than symbolic policies or template-based generation.

\subsection{LLM-Grounded Dialogue State and Move Selection}
\label{sec:controller}

The controller is a lightweight stateful sidecar that reads mean-pooled
residual-stream activations from the frozen backbone. For each context turn
$\tau<t$, it standardizes the layer-$\ell_{\mathrm{ctrl}}^\star$ representation
$x_\tau\in\mathbb{R}^d$ and updates a selective state-space model
\citep{gu2023mamba}:
\begin{equation}
z_\tau=\bar a_\tau\odot z_{\tau-1}+\bar b_\tau,
\qquad
o_\tau=C_\tau z_\tau,
\label{eq:ctrl-rec}
\end{equation}
where $\bar a_\tau,\bar b_\tau$ depend on $x_\tau$ and
$\bar a_\tau\in(0,1)^N$. A linear head predicts the next move, and an auxiliary
head regresses log response length:
\begin{equation}
p_\theta(y_t\mid c_t)=\operatorname{softmax}(W_o o_{t-1}),
\qquad
\widehat L_t=w_L^\top o_{t-1}.
\end{equation}
The length output is monotonically calibrated to a turn-boundary steering
strength and, like all steering decisions, acts through the move-level gating
rule below; move-steering strengths are per-move constants selected on training
data.

\paragraph{Realization-aware gating.}
Based on training-set steerability, move steering is withheld for
\textsc{reply}, which the unsteered backbone realizes more reliably, while
turn-boundary steering is applied only to naturally terse moves
(\textsc{acknowledge}, \textsc{query}, and \textsc{reply}). All gates depend on
the controller prediction rather than gold labels.

\paragraph{Training and inference.}
With the backbone frozen, we train the controller using next-move
cross-entropy and smooth-$L_1$ log-length loss. All components use the same
dialogue-level split, excluding held-out test dialogues. At inference, the
controller maintains a persistent streaming state across turns.

\subsection{Conversational-Move Subspaces for Realization}
\label{sec:vectors}

For each follower turn $t$ in the training dialogues, with gold move
$y_t\in\mathcal{M}$, we extract the residual-stream change at response onset.
Let $h_t^\ell(s)\in\mathbb{R}^d$ denote the activation at layer $\ell$ and token
position $s$, and let $s_{t,1}$ denote the position of the first token of response $u_t$. We define
\begin{equation}
\Delta_t^\ell =h_t^\ell(s_{t,1})-h_t^\ell(s_{t,1}-1),
\label{eq:move-onset}
\end{equation}
which captures the transition from dialogue context to response generation. For each backbone, move directions are estimated exclusively from its activations
on the MapTask training split and then applied, without re-estimation, to the same
backbone on the held-out MapTask test split, FindTask, and CReST.

At each layer, we standardize the onset differences $\{\Delta_t^\ell\}_{t \in \mathcal{I}_{\text{train}}}$, project them into a $k$-dimensional PCA space, and estimate a one-versus-rest Fisher direction for
each move using Ledoit--Wolf shrinkage
\citep{ledoit2004well,fisher1936use}. Let $\nu_m^\ell$ and
$\nu_{\neg m}^\ell$ denote the means of move $m$ and all remaining moves in
PCA space, and let $\widehat{\Sigma}_{w,m}^\ell$ denote the shrinkage estimate
of their pooled within-class covariance. The discriminative direction is
\begin{equation}
w_m^\ell
\propto
\left(\widehat{\Sigma}_{w,m}^\ell\right)^{-1}
\left(\nu_m^\ell-\nu_{\neg m}^\ell\right).
\label{eq:lda}
\end{equation}

We map $w_m^\ell$ back to residual-stream coordinates, undo the feature
scaling, and normalize the result to obtain the move direction $v_m^\ell$.
The inverse covariance suppresses dimensions with high within-class variation,
yielding a more discriminative direction than a simple difference of means.

For target move $m$, we inject its direction at the selected layer
$\ell_m^\star$ at every generated token position $s$:
\begin{equation}
h^{\ell_m^\star}(s)
\leftarrow
h^{\ell_m^\star}(s)
+
\alpha c_{\ell_m^\star}v_m^{\ell_m^\star}.
\label{eq:additive}
\end{equation}
where $c_{\ell_m}$ is the median residual-stream norm at that layer and
$\alpha$ is a dimensionless steering coefficient. This norm-relative scaling
improves comparability across layers and model backbones.

\subsection{Automatic Move Classifier}
\label{sec:move-classifier}

We use Qwen2.5-72B \citep{yang2025qwen3} as a prompted few-shot classifier over five move types, using the same prompt and decoding procedure across datasets. Classification is context-dependent: we first label the preceding giver turn, then classify the follower response conditioned on that label, using the gold giver label when available. Decoding is restricted to one option token, whose normalized log-probabilities define \(P(\text{move}\mid\cdot)\). On human corpus responses, the classifier recovers the gold move with \(77.2\%\)--\(81.6\%\) accuracy (Table~\ref{tab:classifier-validation}). For generation evaluation, we withhold the human response, generate a replacement, and compare its classifier-assigned move with the original human gold label. We further validate it on \(500\) generated responses, balanced across target moves, datasets, and backbone models. Three blinded annotators label each response from its dialogue context. The classifier agrees with their majority label on \(73.9\%\) of examples, while human agreement is Fleiss' \(\kappa=0.682\). Across the seven generation methods, classifier-based and human-majority move accuracies are strongly correlated (Spearman \(\rho=0.96\)) and induce nearly the same ranking, so the automatic metric does not systematically favor any particular method.

\begin{table}[H]
\centering
\small
\setlength{\tabcolsep}{5pt}
\renewcommand{\arraystretch}{1.08}
\begin{tabular}{lccc}
\toprule
\textbf{Dataset}
& \shortstack{\textbf{Corpus}\\\textbf{Agreement}}
& \shortstack{\textbf{Generated}\\\textbf{Agreement}}
& \shortstack{\textbf{Human}\\\textbf{Fleiss' $\kappa$}} \\
\midrule
MapTask  & 78.6\% & 76.4\% & 0.659 \\
FindTask & 81.6\% & 67.8\% & 0.615 \\
CReST    & 77.2\% & 75.5\% & 0.718 \\
\midrule
Overall  & 78.5\% & 73.9\% & 0.682 \\
\bottomrule
\end{tabular}
\caption{Move-classifier validation against corpus labels and majority judgments from three blinded annotators; Fleiss' $\kappa$ measures inter-annotator agreement.}
\label{tab:classifier-validation}
\end{table}

\section{Results}
\label{sec:results}

We evaluate the proposed framework at three levels: next-move selection, oracle-target realization, and end-to-end conversational control. Experiments span Qwen2.5-Omni \citep{xu2025qwen25omnitechnicalreport}, Qwen3-Omni \citep{xu2025qwen3}, and Phi-4-Multimodal \citep{abouelenin2025phi}.

\subsection{Selection: Predicting the Next Move}

\begin{table}[H]
\centering
\small
\setlength{\tabcolsep}{5pt}
\renewcommand{\arraystretch}{1.08}
\resizebox{\columnwidth}{!}{%
\begin{tabular}{lccc@{\hspace{20pt}}c}
\toprule
\textbf{Selector} & \textbf{MapTask} & \textbf{FindTask} & \textbf{CReST} & \textbf{Avg.} \\
\midrule
Majority baseline & 0.47 & 0.41 & 0.37 & 0.42 \\
TF-IDF $+$ logistic (last 5 turns) & 0.54 & 0.46 & 0.44 & 0.48 \\
GPT-4o (few-shot) & 0.40 & 0.32 & 0.53 & 0.42 \\
Qwen2.5-72B (few-shot) & 0.38 & 0.38 & 0.41 & 0.39 \\
Markov-1 (gold prev.\ move) & 0.65 & 0.42 & 0.56 & 0.54 \\
Transcript-embed.\ probe & 0.59 & 0.53 & 0.58 & 0.57 \\
Transcript-embed.\ SSM & 0.58 & 0.50 & 0.60 & 0.56 \\
Activation probe & \textbf{0.64} & 0.50 & \textbf{0.61} & 0.58 \\
\midrule
Controller (ours, streaming) & 0.62 & \textbf{0.58} & \textbf{0.61} & \textbf{0.60} \\
\bottomrule
\end{tabular}
}
\caption{Next-move selection accuracy. Markov-1 receives the gold preceding
move; the activation probe and the controller read the residual stream at the
selected controller layer $\ell_{\mathrm{ctrl}}^\star$.}
\label{tab:selection-results}
\end{table}

 The activation controller outperforms all transcript-based selectors across datasets. On MapTask it reaches \(0.62\), close to the gold-history Markov baseline at \(0.65\), while on FindTask and CReST it surpasses that baseline. The non-recurrent activation probe performs similarly on MapTask and CReST, suggesting that the backbone already integrates the relevant dialogue history. On FindTask, the stateful controller improves over the probe by \(0.08\), indicating that explicit cross-turn state is useful for shorter episodes.

\subsection{Realization: Generating a Target Move}

\begin{table}[h]
\centering
\scriptsize
\setlength{\tabcolsep}{2.5pt}
\renewcommand{\arraystretch}{.84}

\resizebox{\columnwidth}{!}{%
\begin{tabular}{llcccccc}
\toprule
\textbf{Dataset} & \textbf{Model}
& \textbf{Base}
& \textbf{SI}
& \textbf{PBL}
& \textbf{FUDGE}
& \textbf{DeAL}
& \textbf{Ours} \\
\midrule

\multirow{3}{*}{MapTask}
& Qwen3 & 29.0 & 30.1 & \textbf{43.2} & 29.9 & 34.9 & 41.0 \\ 
& Qwen2.5 & 30.9 & 38.8 & 42.4 & 30.7 & 39.8 & \textbf{66.8} \\
& Phi-4 & 40.6 & 63.2 & 56.6 & 46.7 & 31.1 & \textbf{64.8} 
\\ \midrule 

\multirow{3}{*}{FindTask} 
& Qwen3 & 29.5 & 34.6 & 46.2 & 26.9 & 37.8 & \textbf{52.6} \\
& Qwen2.5 & 23.7 & 25.6 & 34.0 & 31.4 & 36.5 & \textbf{55.8} \\
& Phi-4 & 29.5 & 35.3 & 35.9 & 25.6 & 33.3 & \textbf{50.6} \\ 
\midrule 

\multirow{3}{*}{CReST} 
& Qwen3 & 42.9 & 48.3 & \textbf{64.3} & 46.7 & 55.8 & 63.3 \\
& Qwen2.5 & 39.8 & 43.9 & 62.1 & 41.4 & 53.0 & \textbf{66.8} \\
& Phi-4 & 43.6 & 75.2 & 62.1 & 36.7 & 46.4 & \textbf{78.1} \\ 
\midrule 

\multicolumn{2}{l}{\textbf{Avg.}}
& 34.4 & 43.9 & 49.6 & 35.1 & 41.0 & \textbf{60.0} \\ 
\bottomrule
\end{tabular}
}

\caption{Oracle-target move-realization accuracy (\%). All methods receive the
gold target move.}
\label{tab:realization-results}
\end{table}

\begin{table*}[t]
\centering
\scriptsize
\setlength{\tabcolsep}{3.5pt}
\renewcommand{\arraystretch}{.84}

\resizebox{\textwidth}{!}{%
\begin{tabular}{ll*{5}{cc}}
\toprule
\textbf{Dataset}
& \textbf{Model}
& \multicolumn{2}{c}{\textbf{Baseline}}
& \multicolumn{2}{c}{\textbf{Prompt Control}}
& \multicolumn{2}{c}{\textbf{PAS}}
& \multicolumn{2}{c}{\textbf{SFT}}
& \multicolumn{2}{c}{\textbf{\textsc{Latent-IM}}} \\
\cmidrule(lr){3-4}
\cmidrule(lr){5-6}
\cmidrule(lr){7-8}
\cmidrule(lr){9-10}
\cmidrule(lr){11-12}

&
& Acc. & BLEU
& Acc. & BLEU
& Acc. & BLEU
& Acc. & BLEU
& Acc. & BLEU \\
\midrule

\multirow{3}{*}{MapTask}
& Qwen3
& 28.98 & 8.38
& 19.39 & 16.66
& 25.25 & 8.72
& \textbf{45.05} & 26.44
& 30.35 & 10.12 \\

& Qwen2.5
& 30.87 & 11.29
& 24.24 & 10.24
& 29.90 & 10.50
& \textbf{46.46} & 26.29
& 41.33 & 20.52 \\

& Phi-4
& 40.59 & 19.68
& 33.13 & 14.56
& 34.75 & 14.58
& 46.26 & 25.66
& \textbf{46.47} & 20.33 \\

\midrule

\multirow{3}{*}{FindTask}
& Qwen3
& 29.49 & 8.01
& 39.10 & 16.43
& 27.56 & 8.10
& 44.87 & 26.56
& \textbf{49.36} & 10.61 \\

& Qwen2.5
& 23.72 & 12.24
& 34.62 & 15.07
& 21.79 & 11.83
& 25.00 & 17.30
& \textbf{48.72} & 16.34 \\

& Phi-4
& 29.49 & 6.96
& 28.95 & 10.68
& 34.62 & 6.63
& 45.51 & 31.69
& \textbf{48.08} & 6.92 \\

\midrule

\multirow{3}{*}{CReST}
& Qwen3
& 42.95 & 9.49
& 12.54 & 19.29
& 45.77 & 10.17
& \textbf{56.11} & 23.79
& 52.04 & 10.25 \\

& Qwen2.5
& 39.81 & 12.62
& 21.00 & 10.81
& 42.01 & 12.94
& \textbf{55.49} & 23.87
& 47.02 & 16.61 \\

& Phi-4
& 43.57 & 17.59
& 35.42 & 10.60
& 46.08 & 14.22
& 54.86 & 24.06
& \textbf{58.62} & 18.14 \\

\bottomrule
\end{tabular}%
}

\caption{
End-to-end move-control results across dialogue datasets and multimodal
language models. Acc.\ is the percentage of generated responses whose predicted
conversational move matches the held-out human move. BLEU measures lexical
overlap with the held-out human response.
}
\label{tab:main-results}
\end{table*}

We evaluate oracle realization by providing every method with the dialogue context and gold follower move. Baselines include direct prompting (Base), few-shot candidate selection (PBL) \citep{ramirez2023controllablegenerationdialogueacts}, instruction-derived steering (SI) \citep{stolfo2025improvinginstructionfollowinglanguagemodels}, classifier-guided decoding (FUDGE) \citep{Yang_2021}, and reward-guided search (DeAL) \citep{Huang_2025}. \textsc{Latent-IM} applies move directions learned exclusively from MapTask and transferred unchanged to all evaluation sets. It achieves the highest mean accuracy (\(60.0\%\)) and leads in six of nine settings.

\subsection{End-to-End Move Control}

Finally, we evaluate end-to-end response generation, where all methods receive the same held-out dialogue context and no gold move labels at inference. Table~\ref{tab:main-results} compares two explicit selection--realization pipelines with three direct-generation baselines. Prompt Control first predicts a conversational move using an LLM selector and then includes that prediction in a definitional prompt. \textsc{Latent-IM} instead uses the streaming activation controller to select the next move and realizes it through fixed move-specific steering directions. Base, PAS, and SFT do not explicitly predict or condition on move labels: Base generates directly from the dialogue context, PAS applies its learned intervention directly during generation, and SFT separately fine-tunes each backbone with LoRA \citep{hu2022lora} on context--response pairs to imitate the human response.

\textsc{Latent-IM} achieves the highest mean end-to-end move accuracy at \(46.9\%\), slightly exceeding SFT at \(46.6\%\) and outperforming Base (\(34.4\%\)), PAS (\(34.2\%\)), and Prompt Control (\(27.6\%\)). It obtains the highest accuracy in the majority of the dataset--model settings. SFT remains stronger for Qwen3 and Qwen2.5 on MapTask and CReST and produces higher BLEU \citep{papineni2002bleu} scores overall, consistent with direct supervised imitation of the reference response. In contrast, \textsc{Latent-IM} often realizes the appropriate conversational move using lexically different wording. These results show that explicitly recovering and controlling an intermediate conversational action can match direct supervised fine-tuning while retaining a modular, label-free inference pipeline.

\section{Turn-Boundary Subspace}
\label{sec:turn-boundary}

\begin{figure}
    \centering
    \includegraphics[width=0.95\linewidth]{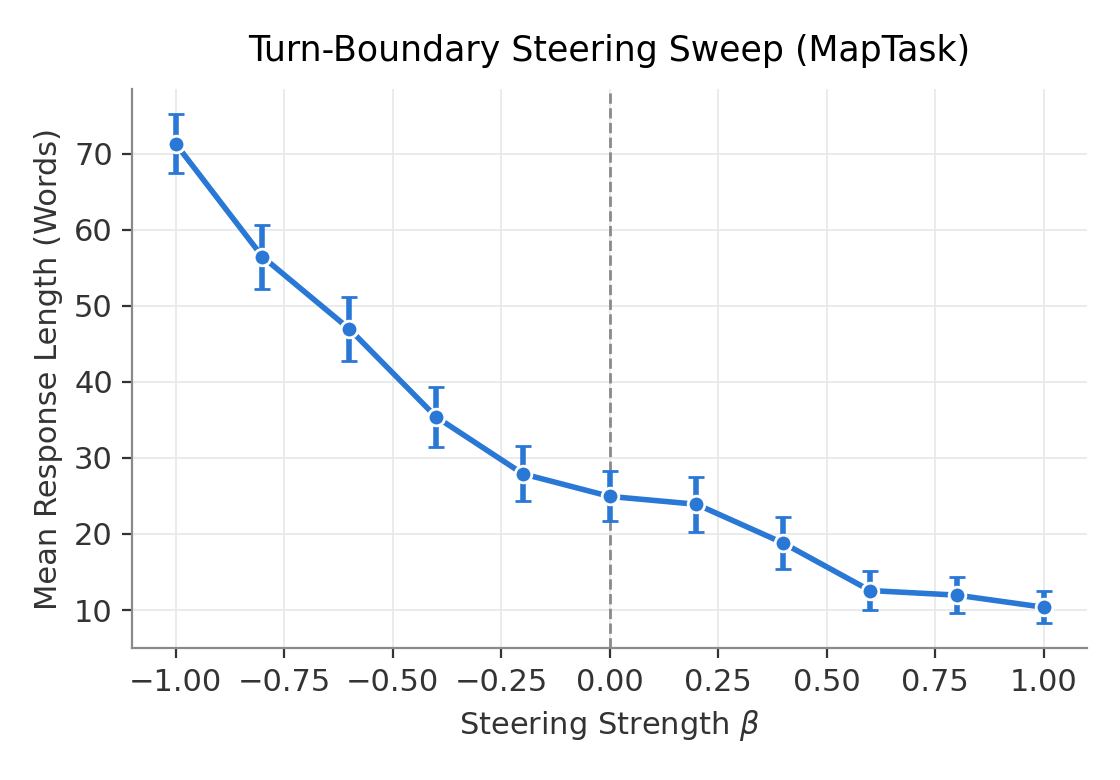}
    \caption{Steering along the turn-boundary direction monotonically controls
    response length (mean $\pm$95\% CI).}
    \label{fig:eot-sweep}
\end{figure}

Predicting whether a speaker will continue or yield the floor is a longstanding problem in spoken dialogue modeling \citep{skantze2017towards,ekstedt2020turngpt,lin2025predicting}. We identify a turn-boundary direction representing whether the current response should continue or terminate, complementing acoustic pauses that may occur within a turn \citep{lala2017attentive}. We derive supervision from the speaker sequence of the MapTask training split. For each turn \(t<T\), we assign
\begin{equation}
b_t=\mathbf{1}\!\left[r_{t+1}\neq r_t\right],
\end{equation}
where \(b_t=1\) denotes yielding the floor and \(b_t=0\) denotes continuing. At each layer, we mean-pool the residual stream over the turn's token span. A layer sweep selects layer $\ell^\star_{\textsc{eot}}$, and the unit-norm turn-boundary direction is estimated by the difference between the yield and hold class means:
\begin{equation}
v_{\textsc{eot}}
=
\frac{\mu_{\textsc{yield}}-\mu_{\textsc{hold}}}
{\lVert\mu_{\textsc{yield}}-\mu_{\textsc{hold}}\rVert_2}.
\end{equation}
Like the move directions, \(v_{\textsc{eot}}\) is learned only from MapTask training dialogues and transferred without re-estimation to all evaluation sets. The direction is predictive: on held-out follower turns it detects upcoming boundaries at \(0.95\) AUROC from content alone, versus \(0.932\) for a pause-only detector, which is also post hoc since it can only score a boundary after the silence has elapsed. An incremental spoken system needs the former to yield the floor without a gap. During generation, we inject the direction using the same norm-relative rule as move steering:
\begin{equation}
h^{\ell^\star_{\textsc{eot}}}(s)
\leftarrow
h^{\ell^\star_{\textsc{eot}}}(s)
+
\beta c_{\ell^\star_{\textsc{eot}}}v_{\textsc{eot}}.
\end{equation}
Positive \(\beta\) encourages termination, while negative \(\beta\) prolongs the response. Figure~\ref{fig:eot-sweep} sweeps \(\beta\) over \([-1,1]\) on \(250\) held-out test contexts with only the turn-boundary direction injected: mean response length varies monotonically from \(71.3\) words at \(\beta{=}-1\) through the unsteered \(24.9\) down to \(10.4\) at \(\beta{=}1\), with decoding remaining fluent throughout, giving graded, sample-free control over when the model yields the turn. 

\section{Analysis}

\subsection{Controller}

{\paragraph{State is read, not recomputed.} Table~\ref{tab:selection-results} separates
what carries the dialogue state from what reads it. The transcript-embed rows see
the same token sequence as the controller but through the backbone's embedding
table alone; with matched readers, replacing contextual activations with these
lexical features costs three to five points on MapTask and CReST, and on FindTask
the gap is closed only by a reader that carries explicit cross-turn state. The
state is computed by the frozen backbone and merely read out. Reading it is also
more reliable than asking for it: prompting GPT-4o or Qwen2.5-72B to name the next
move falls $0.08$--$0.26$ below the 0.5M controller. Finally, the reader's
architecture barely matters: MLP, GRU, LSTM, and SSM variants all reach
$\approx0.60$ on MapTask, which is what one expects if the state is linearly
available in the residual stream at each turn. We adopt the selective SSM because
it formalizes the per-turn state update with a constant-time streaming step, not
because recurrence wins on accuracy; the exception is FindTask's short episodes,
where stateless readers drop to $\approx0.50$ and carried state is what remains.}

\paragraph{Selection is not enough.} Higher selection accuracy does not always
translate into better end-to-end control, because per-move predictability and
steerability tend to trade off: a move the controller predicts confidently is not
necessarily one that steering can realize. The clearest case is \textsc{reply}:
after a giver question the follower almost always replies, so a giver-move prior
sharply improves \textsc{reply} selection (precision $0.53\!\to\!0.64$), yet this
yields no end-to-end gain, because the \textsc{reply} steering direction
surface-overlaps \textsc{acknowledge} and does not land (controllability $0.14$ vs.\
$0.99$ for \textsc{query}). This motivates \emph{realization-aware gating}: apply the
action-subspace steering only where it lands.

\paragraph{Selection under ambiguity.}
Selection accuracy must be interpreted against how consistent \emph{humans}
are in the same situation. Even near-identical giver turns license different
follower moves: in our held-out set, after the identical giver turn
``\emph{do you see what I mean}'' one follower \textsc{replies}
(``yeah'') while another \textsc{queries} (``so do I go up straight first?'');
after near-identical instructions (``\emph{\ldots centimetres from the
top/bottom of the page}'') one follower \textsc{checks} (``the bottom of the
page you mean'') while another \textsc{acknowledges} (``mm right''). We find
55 such similar-giver/different-follower pairs in the test dialogues alone.
Aggregating by the preceding giver move, the gold follower move is
concentrated after questions (after \textsc{query-yn}, $75\%$ of humans reply)
but genuinely open after instructions and acknowledgements (mode probability
$0.30$--$0.51$).

The controller inherits exactly this structure. Splitting held-out turns by
human consistency: on \emph{clear} contexts (mode $\geq0.6$) it reaches $0.61$
against a human-consistency ceiling of $0.68$ ($89\%$); on \emph{open} contexts
it reaches $0.46$ against a ceiling of $0.49$ ($95\%$). Per giver-move type,
controller accuracy tracks the human mode probability nearly one-to-one
(e.g., \textsc{query-yn}: $0.758$ vs.\ $0.75$), and its predictive entropy
correlates with the true follower-move entropy at $r=0.95$. Its confusions
flow into \textsc{acknowledge}---the modal move---which is the Bayes-optimal
guess under ambiguity rather than a failure mode. The ceiling is also
modality-invariant: adding the follower's map as text ($0.526$) or as an
image through the vision pathway ($0.521$) leaves selection at the no-map
level ($0.524$). Together these results indicate the remaining selection gap
is irreducible follower choice, not missing features; moving past it requires
scoring moves by task outcome rather than matching a single human reference,
which we leave to future work.

\paragraph{Realization-aware gating.} The remedy is a gating rule on the action
subspace, not a new controller: for each predicted move we decide whether to inject
the move direction and whether to inject the turn-length direction, both read from
the controller's own prediction (no gold labels). The rule follows measured per-move
steerability: steer where the controllability matrix is strong (\textsc{check},
\textsc{query}, \textsc{explain}), defer where it is weak (\textsc{reply}). Deferring
\textsc{reply} raises end-to-end realization from $0.67$ to $0.75$ at no selection
cost; length-aware control adds a further gain to $0.78$. {Length is handled as
move policy rather than per-turn regression: the controller maps its selected move to
a turn-boundary coefficient, pulling the naturally terse moves (\textsc{acknowledge},
\textsc{query}, \textsc{reply}) from $\sim\!25$ words toward the human median of
$2$--$3$.} For content-bearing moves (\textsc{check}, \textsc{explain}) the
intervention is \emph{withheld}: the injection needed to shorten them overwhelms the
move direction and collapses realization (\textsc{explain} $0.76\!\to\!0.28$). A
controller that models the geometry of its own steering outperforms one that only
predicts moves accurately.

\subsection{Steering}
\begin{figure}
    \centering
    \includegraphics[width=0.95\linewidth]{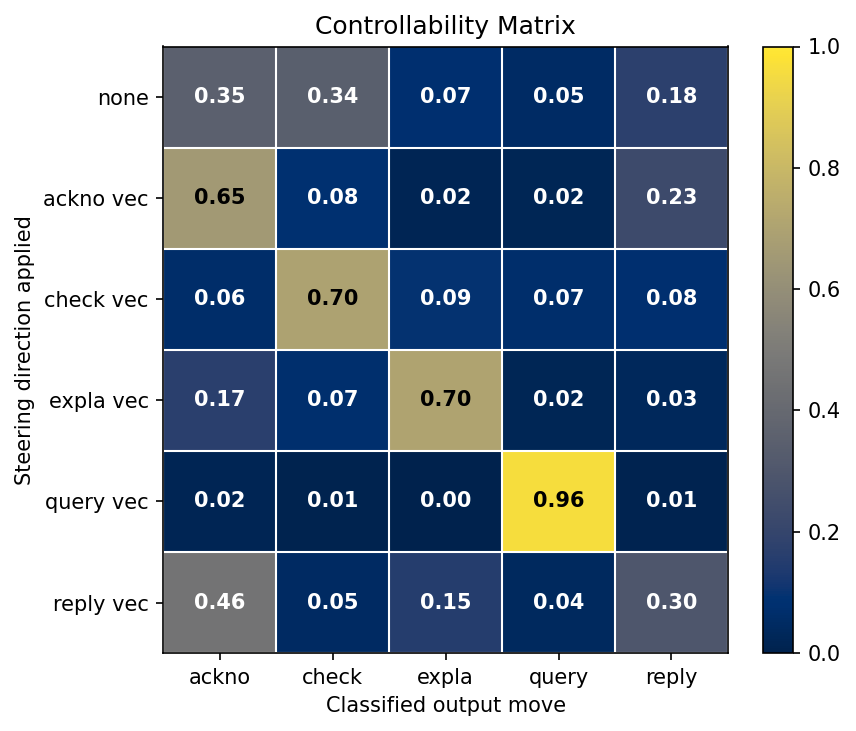}
    \caption{Controllability matrix for move steering (map-task, Qwen2.5-Omni-7B). Each row applies one steering direction to the same held-out contexts; columns give the move assigned to the generation by the classifier}
    \label{fig:controllability_matrix}
\end{figure}

To test whether steering directions are move-specific, we apply each move vector to the same held-out contexts and classify the resulting generations. The resulting controllability matrix shows whether each intervention selectively increases its corresponding conversational move rather than merely perturbing generation. Figure~\ref{fig:controllability_matrix} shows that each steering direction selectively increases its intended move, with a mean diagonal controllability of $0.66$ and limited off-diagonal leakage. Relative to the unsteered baseline, steering raises \textit{query} from $0.05$ to $0.96$ and \textit{explain} from $0.07$ to $0.70$, while \textit{acknowledge} and \textit{check} rise to $0.65$ and $0.70$. These results indicate that the interventions causally target move-specific behavior rather than generically perturbing generation.

Table~\ref{tab:steering-ablations} ablates the main design choices in the steering intervention using oracle target moves; each row varies one component while holding the others at the default configuration. Steering raises target-move accuracy from 0.33 to 0.65, a 0.31-point gain. The modality comparison shows that spoken information contributes signal beyond the transcript alone: audio+text outperforms text-only (0.65 vs. 0.58). The strength sweep exposes a clear controllability tradeoff: accuracy improves up to a moderate strength \((\alpha\approx1.0\), 0.64) but falls to 0.55 at \(\alpha=2.0\). The layer ablation shows that early-layer injection is substantially weaker (0.25) than middle- or late-layer injection (0.48/0.47), while no fixed layer matches the per-move tuned default (0.65). Finally, control is largely insensitive to injection site: residual-stream, attention, and MLP interventions obtain 0.65, 0.65, and 0.64, respectively. Overall, steering depends on multimodal context, moderate intervention strength, and layer choice, while remaining robust to injection location.

\begin{table}[h]
\centering
\footnotesize
\setlength{\tabcolsep}{3pt}
\renewcommand{\arraystretch}{1.08}

\begin{tabular}{@{}p{1.15cm}p{2.35cm}ccc@{}}
\toprule
\textbf{Abl.}
& \textbf{Setting}
& \textbf{Acc.}
& \textbf{$\Delta$}
& \textbf{BLEU} \\
\midrule

Default
& Audio+text, $\alpha^{\star}$, $\ell^\star$, residual
& \textbf{0.65} & -- & 0.18 \\

\midrule
Modality
& Text only
& 0.58 & $-0.07$ & 0.19 \\

\midrule
Strength
& $\alpha=0.0$
& 0.33 & $-0.31$ & 0.11 \\

Strength
& $\alpha=0.5$
& 0.53 & $-0.12$ & 0.14 \\

Strength
& $\alpha=1.0$
& 0.64 & $-0.01$ & 0.17 \\

Strength
& $\alpha=2.0$
& 0.55 & $-0.10$ & 0.19 \\

\midrule
Layer
& Early
& 0.25 & $-0.40$ & 0.10 \\

Layer
& Middle
& 0.48 & $-0.17$ & 0.19 \\

Layer
& Late
& 0.47 & $-0.18$ & 0.09 \\

\midrule
Site
& Attention
& 0.65 & $0.00$ & 0.18 \\

Site
& MLP
& 0.64 & $-0.01$ & 0.18 \\

\bottomrule
\end{tabular}

\caption{Steering ablations on held-out MapTask contexts using
Qwen2.5-Omni-7B and oracle target moves. Each row varies one component of the
default configuration while holding the others fixed. $\Delta$ denotes the
change in target-move accuracy relative to the default; BLEU measures lexical
overlap with the human response.}
\label{tab:steering-ablations}
\end{table}

\subsection{Future Work: Beyond Human-Move Imitation}

Our controller imitates the move a human is likely to make, but the best
task-oriented move may differ from the human choice. Because our steering
mechanism can realize any move in the taxonomy, a future goal-conditioned
controller could select moves according to predicted task utility rather than
imitation likelihood.

\vspace{6pt}
\begin{tcolorbox}[
  colback=gray!4,
  colframe=gray!45,
  title=\textbf{MapTask: an acknowledgment hides a map mismatch},
  fonttitle=\footnotesize,
  left=4pt,
  right=4pt,
  top=4pt,
  bottom=4pt
]
\footnotesize

\noindent\textbf{Giver:}
``Walk north until you reach the Roman baths.''
\textit{[instruct]}

\medskip
\noindent
\colorbox{yellow!18}{%
\parbox{\dimexpr\linewidth-2\fboxsep\relax}{%
\textbf{Follower:} ``Uh-huh.'' \textit{[acknowledge]}%
}}
\medskip

\noindent\textbf{Giver:}
``Then walk right around them.''
\textit{[instruct]}

\noindent\textbf{Follower:}
``Roman baths? I don't have any Roman baths.''
\textit{[query]}

\medskip
The follower initially acknowledges an instruction whose referent is absent
from their map. Although acknowledgment matches the observed human move, an
earlier \textbf{check} or \textbf{query}, such as ``Roman baths? I do not have
those on my map,'' would reveal the mismatch before it propagates into later
instructions.
\end{tcolorbox}

\section{Conclusion}
We introduced \textsc{Latent-IM}, which separates conversational-move selection from realization within a frozen speech LLM. A streaming controller reads residual-stream activations to select the next move, reaching 0.60 average accuracy, while activation steering improves oracle realization by 10.4 percentage points over the strongest baseline. A complementary turn-boundary direction reduces response length by 9.1 words, and the full system improves end-to-end move accuracy by 12.5 points over the unsteered backbone while performing comparably to fine-tuning. These results provide evidence that frozen speech LLM activations can serve as an observable and causally controllable interface for local interaction management. Future work can replace imitation-based selection with goal-conditioned policies that choose moves according to predicted task utility.


\clearpage
\bibliography{aaai2027}

\clearpage
\appendix 
\label{sec:appendix}
\input{appendix}



\end{document}

%% file: appendix.tex
\newcommand{\slot}[1]{\texttt{\{#1\}}}

\tcbset{
  promptbox/.style={
    colback=gray!3,
    colframe=gray!40,
    boxrule=0.4pt,
    arc=2pt,
    left=4pt,
    right=4pt,
    top=4pt,
    bottom=4pt,
    before skip=6pt,
    after skip=6pt,
    breakable,
    enhanced
  }
}

\section{Implementation Details and Prompts}
\label{app:implementation}

\subsection{Generation Prompts}

We use the same prompt structure across datasets, with dataset-specific role and
task descriptions. The placeholder \slot{CONTEXT} contains the preceding
dialogue transcript, while \slot{CONTEXT\_AUDIO} contains the aligned context
audio when available.

\paragraph{MapTask.}\mbox{}\par
\vspace{1pt}

\begin{tcolorbox}[promptbox]
\textbf{System:}
You are the follower in a collaborative MapTask dialogue. Your partner, the
giver, describes a route on a map. Follow their instructions and ask for
clarification when needed.

\tcblower

\textbf{User:}
\texttt{[AUDIO: \{CONTEXT\_AUDIO\}]}
\slot{CONTEXT}
\end{tcolorbox}

\paragraph{FindTask.}\mbox{}\par
\vspace{1pt}

\begin{tcolorbox}[promptbox]
\textbf{System:}
You are the helper in a collaborative object-finding dialogue. Your partner,
the elder, directs you to locate objects in a kitchen. Follow their instructions
and ask for clarification when needed.

\tcblower

\textbf{User:}
\slot{CONTEXT}
\end{tcolorbox}

\paragraph{CReST.}\mbox{}\par
\vspace{1pt}

\begin{tcolorbox}[promptbox]
\textbf{System:}
You are the searcher in a collaborative remote-search dialogue. Your partner,
the director, guides you through a series of rooms to locate numbered boxes.
Follow their directions and ask for clarification when needed.

\tcblower

\textbf{User:}
\texttt{[AUDIO: \{CONTEXT\_AUDIO\}]}
\slot{CONTEXT}
\end{tcolorbox}

\subsection{Automatic Move Classifier}
\label{app:classifier-prompt}

The automatic evaluator classifies the generated follower turn into one of five
moves:
\[
\mathcal{M}
=
\{\texttt{acknowledge},\texttt{check},\texttt{explain},
\texttt{query},\texttt{reply}\}.
\]
Decoding is constrained to these five label tokens. To remove
dataset-specific cues, speaker prefixes are normalized to \texttt{A} and
\texttt{B}. The classifier receives the immediately preceding turn, its move
label when available, and the follower response being evaluated.

\begin{tcolorbox}[promptbox]
\textbf{System:}

Two people are doing a task together by talking. Read the recent exchange and
classify the FINAL turn by what it is doing in the conversation. Choose exactly
one:

\medskip

\texttt{acknowledge} -- only signals that the speaker heard or registered the
other; adds no new information and asks nothing (``okay,'' ``mm-hmm,''
``right,'' ``uh-huh,'' ``I see'').

\smallskip

\texttt{check} -- confirms something already said or inferred by restating,
paraphrasing, or pointing to it (``so I go left?'', ``the blue one?'',
``this one?'').

\smallskip

\texttt{explain} -- volunteers information about the speaker's own situation,
view, or what they can or cannot see that the other did not ask for
(``I don't have that,'' ``there's a box in front of me'').

\smallskip

\texttt{query} -- asks for new information the other has not yet given
(``where do I go?'', ``how many are there?'').

\smallskip

\texttt{reply} -- answers a question the other speaker just asked
(``yes,'' ``no,'' ``it's on the left'').

\medskip

Use the preceding turn to decide. If it asked a question or made a check, the
final turn is normally a \texttt{reply}, even when it contains only one word
such as ``yes,'' ``no,'' or ``okay.'' An exception occurs when the final speaker
ignores the question and instead asks their own question or volunteers
information about their situation.

If the preceding turn was an instruction or statement, the final turn is not a
\texttt{reply}; classify it as an \texttt{acknowledge}, \texttt{check},
\texttt{query}, or \texttt{explain} based on its content.

A short token such as ``yeah,'' ``right,'' or ``okay'' is a \texttt{reply} when
it answers a question, but an \texttt{acknowledge} when it merely registers an
instruction or statement.

A turn can be a \texttt{check} even without explicit question syntax when it
repeats or reformulates specific preceding information to confirm it, as in
``to the right,'' ``past the door,'' or ``the blue box then.'' A single bare
word should not be treated as a check.
\end{tcolorbox}

\begin{tcolorbox}[promptbox]
\textbf{User:}

Recent exchange:

\slot{CONTEXT}

\medskip

The preceding turn was labeled:
\slot{PRECEDING\_MOVE}

\medskip

Final turn to classify:

``\slot{MODEL\_RESPONSE}''

\medskip

Answer with exactly one word from:
\texttt{acknowledge}, \texttt{check}, \texttt{explain},
\texttt{query}, or \texttt{reply}.

\medskip

\textbf{Answer:}
\end{tcolorbox}

\subsection{Baselines}
\label{app:baselines}

Prompt-Based Learning, Steering-Instruct, and FUDGE receive the target move in
the oracle-realization setting. SFT is an end-to-end baseline and receives no
move label during training or inference.

\paragraph{Prompt-Based Learning (PBL).}

PBL \citep{ramirez2023controllablegenerationdialogueacts} prepends \(N=10\) in-context demonstrations sampled from the training split
using seed \(42\). Each demonstration contains at most the three most recent
context turns. Demonstrations are sampled at run time rather than drawn from a
single fixed set.

The default target-shaped template is:

\begin{tcolorbox}[promptbox]
\texttt{Here is a dialogue context:}

\slot{CONTEXT}

\medskip

\texttt{The follower's next response, written as a
`\slot{MOVE}' dialogue act: \slot{RESPONSE}}
\end{tcolorbox}

At test time, the same template is used but ends immediately after
\texttt{dialogue act:}. The \texttt{definitional} variant prepends the target
move's definition. The \texttt{fewshot} variant instead uses:

\begin{tcolorbox}[promptbox]
\texttt{Context:} \slot{CONTEXT}

\medskip

\texttt{Response (\slot{MOVE}):} \slot{RESPONSE}
\end{tcolorbox}

For each example, the model samples \(K=10\) candidates using temperature
\(0.7\), top-\(p=1.0\), and at most \(60\) new tokens. Among candidates that the
move classifier assigns to the target move, PBL selects the one with the highest
mean token log-probability. If no candidate is assigned to the target move, it
selects
\[
\hat S
=
\arg\max_S
P_{\mathrm{cls}}(m\mid c_t,S)\,P_G(S\mid c_t).
\]

\paragraph{Steering-Instruct.}

Steering-Instruct \citep{stolfo2025improvinginstructionfollowinglanguagemodels} contrasts a bare dialogue context against three
move-conditioned prompt variants:

\begin{tcolorbox}[promptbox]
\texttt{Your next response should \slot{DEFINITION}.}

\smallskip

\texttt{Respond in a way that will \slot{DEFINITION}.}

\smallskip

\texttt{Make sure your reply will \slot{DEFINITION}.}
\end{tcolorbox}

Using \(200\) training contexts sampled with seed \(42\), we collect the
last-input-token residual at every layer for both the base and instructed
prompts. At layer \(\ell\), the direction is
\[
v_m^\ell
=
\mu_{\mathrm{instr},m}^\ell-\mu_{\mathrm{base}}^\ell,
\qquad
u_m^\ell
=
\frac{v_m^\ell}{\lVert v_m^\ell\rVert_2}.
\]
We also store the target projection
\[
\bar z_m^\ell
=
\left\langle
\mu_{\mathrm{instr},m}^\ell,
u_m^\ell
\right\rangle.
\]

During generation, a projection-matching hook is applied at
\(\ell=\lfloor L/2\rfloor\). At each generated token, the intervention
coefficient is recomputed so that the residual's projection onto
\(u_m^\ell\) matches \(\bar z_m^\ell\). We use gain \(1.0\) and greedy decoding.
The move instruction is used only to construct the direction and is not included
in the generation prompt.

\paragraph{FUDGE.}

FUDGE \citep{Yang_2021} uses a lightweight future discriminator consisting of a
\(128\)-dimensional token embedding, a single-layer LSTM with hidden dimension
\(256\), and a linear output layer over the five moves. The discriminator is
trained on response prefixes from the training dialogues to predict the
completed response's move. We apply class-balanced cross-entropy at every prefix
position and train using Adam with learning rate \(10^{-3}\), batch size \(64\),
and \(8\) epochs. Ten percent of the training dialogues are reserved for
validation, and the checkpoint with the highest macro accuracy is retained.

At decoding step \(i\), the generator's top \(K=200\) tokens are scored as
\[
\operatorname{score}(x_i)
=
\log P_G(x_i\mid x_{<i},c_t)
+
\lambda
\log P_B(m\mid x_{\leq i},c_t),
\]
where \(\lambda=1\). The highest-scoring token is selected greedily, without
sampling or beam search.

\paragraph{Supervised Fine-Tuning (SFT).}

SFT is an end-to-end baseline that fine-tunes the backbone on context--response
pairs,
\[
c_t\longrightarrow u_t,
\]
without providing a move label during training or inference \citep{ouyang2022training}.

We apply LoRA with rank \(r=16\), scaling parameter \(32\), dropout \(0.05\), and all linear layers as target modules \citep{hu2022lora}. Training is text-only and uses up to \(1{,}000\) training pairs, with loss computed only over response tokens. We use AdamW with learning rate \(2\times10^{-4}\), three epochs, per-device batch size \(1\), gradient accumulation over \(16\) steps, BF16 precision, and seed \(42\). The final-epoch weights are retained without validation-based checkpoint selection. For Qwen2.5-Omni, approximately \(65\)M of \(9.0\)B parameters are trainable, corresponding to \(0.72\%\).

\paragraph{DeAL.}

DeAL \citep{Huang_2025} is a decoding-time alignment baseline that performs an $A^{\ast}$-style search---top-$k$ expansion, greedy lookahead, and a self-reward heuristic---toward the target move, without modifying weights or activations. It is text-only. Optionally the context is augmented with an alignment instruction $p_a$,

\begin{tcolorbox}[promptbox]
\slot{CONTEXT}

\medskip

\texttt{Respond with a \slot{MOVE} move: \slot{DEFINITION}}
\end{tcolorbox}

\noindent but in our main runs this instruction is dropped
(\texttt{--no-align-prompt}) so that alignment is driven purely by the search
reward and does not double-count the move-classifier signal. At each committed
step we (i) take the generator's top $K=5$ next-token candidates with their
log-probabilities, (ii) extend each candidate with a greedy lookahead of
$L=16$ tokens, and (iii) score a self-reward
\[
h_j
=
\log P_{G}\!\left(m \mid c_t,\, \text{committed}\oplus\text{cand}_j\oplus\text{lookahead}\right),
\]
read from the \emph{same} generation model via A--E letter log-probabilities
over the five moves. We commit
\[
\hat\jmath
=
\arg\max_{j}\ \bigl[\log P_G(\text{cand}_j) + \lambda\, h_j\bigr],
\]
advancing $\text{stride}=4$ tokens per decision, with $\lambda=0.5$ and at most
$40$ new tokens. Decoding is greedy; there is no sampling or beam search.

\paragraph{PAS.}

PAS (Painless Activation Steering) is a self-supervised, end-to-end baseline that builds a corrective steering vector without any external supervision \citep{cui2026painlessactivationsteeringautomated}. We run the unmodified generation model on the training split's next-move-prediction task,

\begin{tcolorbox}[promptbox]
\texttt{The follower's possible moves are:}\\
\texttt{- acknowledge: \slot{DEFINITION}}\\
\texttt{\ldots}\\
\texttt{- reply: \slot{DEFINITION}}

\medskip

\texttt{Dialogue so far:} \slot{CONTEXT}

\medskip

\texttt{The follower (F) speaks next. Which single move type will it be?}
\texttt{Answer with exactly one of: acknowledge, check, explain, query, reply.}
\texttt{Answer:}
\end{tcolorbox}

\noindent and partition up to $1{,}000$ training contexts (seed $42$) into
correct and incorrect by the model's \emph{own} parsed answer versus the gold
move. The vector at layer $\ell$ is the raw difference of mean last-token
residuals between a positive and a negative prompt set, captured at all layers,
\[
a^{\ast}_{\ell}
=
\operatorname*{mean}_{P_{+}}\!\bigl(h_{\ell},\ \text{last token}\bigr)
-
\operatorname*{mean}_{P_{-}}\!\bigl(h_{\ell},\ \text{last token}\bigr).
\]
We use three variants: \textbf{PASf}, whose $P_{+}/P_{-}$ are the full multiple-choice prompts for correctly/incorrectly answered contexts; \textbf{iPASa}, whose prompts append the model's chosen move and are partitioned by correctness; and \textbf{iPASwo}, built from incorrect contexts only, with $P_{+}$ appending the correct move and $P_{-}$ the chosen wrong move (the default). At generation the vector is added additively at a single layer $\ell=\lfloor L/2\rfloor$ with strength $\lambda=1.0$, using greedy decoding.

\section{Compute}

\begin{table}[ht]
    \centering
    \begin{tabularx}{\columnwidth}{Xc}
        \toprule
        \textbf{Parameter} & \textbf{Value} \\
        \midrule
        temperature & 0.0 \\
        num\_beams & 1 \\
        top\_k & disabled ($-1$) \\
        top\_p & 1.0 \\
        repetition\_penalty & 1.0 \\
        max\_new\_tokens & 8192 \\
        stop\_sequences & none \\
        \bottomrule
    \end{tabularx}
    \caption{Hyperparameters used for decoding.}
    \label{tab:hyperparameters}
\end{table}
We use vLLM \cite{kwon2023efficient} for inference. Parameters are shown in Table \ref{tab:hyperparameters}. All inference was run on Nvidia A40 GPUs with 48GB GDDR6 memory. We use vLLM tensor parallelism across 2, 4, or 8 GPUs depending on model size. Runtime ranged from a few minutes to about one hour per (model, dataset) evaluation, depending on model size and dataset.

\subsection{Models}
\label{app:models}
We evaluate three instruction-tuned multimodal language models: Qwen2.5-Omni, Qwen3-Omni, and Phi-4-Multimodal, using the following Hugging Face checkpoints without modification \cite{wolf2020transformers}.

\begin{table}[h]
\centering
\small
\setlength{\tabcolsep}{3pt}
\begin{tabular}{p{1.25cm} p{3.25cm} p{2.15cm}}
\toprule
\textbf{Family} & \textbf{Hugging Face Model ID} & \textbf{Special tokens} \\
\midrule
Qwen2.5-Omni
&
\texttt{Qwen/Qwen2.5-}
\texttt{Omni-7B}
&
\texttt{<|im\_start|>}, \texttt{<|im\_end|>}, \texttt{<eos>} \\
\midrule
Qwen3-Omni
&
\texttt{Qwen/Qwen3-Omni-}
\texttt{30B-A3B-Instruct}
&
\texttt{<|im\_start|>}, \texttt{<|im\_end|>}, \texttt{<eos>} \\
\midrule
Phi-4
&
\texttt{microsoft/Phi-4-}
\texttt{multimodal-instruct}
&
\texttt{<|user|>}, \texttt{<|assistant|>}, \texttt{<|end|>} \\
\bottomrule
\end{tabular}
\caption{Instruction-tuned models evaluated, with special tokens used for prompt construction.}
\label{tab:models_and_special_tokens}
\end{table}

\section{Datasets and Preprocessing}
\label{app:data}

\subsection{Corpus Statistics and Splits}
We evaluate on three task-oriented dialogue corpora in which one participant issues instructions---the \emph{giver}, \emph{elder}, or \emph{director}---and the other executes them---the \emph{follower}, \emph{helper}, or \emph{searcher}. We normalize these roles to \emph{giver} and \emph{follower}. The follower turn is the unit of prediction and steering, and each retained follower turn constitutes one example.

\begin{table}[h]
\centering
\footnotesize
\setlength{\tabcolsep}{3pt}
\renewcommand{\arraystretch}{1.08}
\begin{tabular}{@{}lrrrc@{}}
\toprule
\textbf{Dataset}
& \textbf{Dlg.}
& \textbf{Turns}
& \textbf{Train/Test}
& \textbf{Input} \\
\midrule
MapTask  & 128 & 11{,}581 & 8{,}876 / 2{,}705 & Audio+text \\
FindTask & 133 & 642      & 486 / 156         & Text \\
CReST    & 17  & 2{,}005  & 1{,}686 / 319     & Audio+text \\
\bottomrule
\end{tabular}
\caption{Dataset statistics after label mapping and filtering. Splits are grouped by dialogue. ``Turns'' counts retained follower turns whose labels map to one of the five target moves.}
\label{tab:data}
\end{table}

All experiments use a single canonical dialogue-grouped split. We hold out \(20\%\) of dialogues for testing, ensuring that no dialogue appears in more than one split. Dialogue identifiers are sorted and sampled without replacement using \texttt{numpy.default\_rng(42)}, with
\[
n_{\mathrm{test}}
=
\max\!\left(
1,
\left\lfloor 0.2|\mathcal{D}|\right\rfloor
\right).
\]
This produces \(25\) MapTask test dialogues, \(26\) FindTask test dialogues,
and \(3\) CReST test dialogues.

Steering-vector estimation, controller training, baseline training, and end-to-end evaluation all use this split. No separate development set is reserved for the main experiments. When an auxiliary validation set is needed, we sample \(10\%\) of the training dialogues: seed \(42\) for the FUDGE discriminator and seed \(123\) for the controller.

\subsection{Move-Label Mapping}

We map corpus-specific dialogue-act labels into the shared move set
\[
\mathcal{M}
=
\{\textsc{acknowledge},\textsc{check},\textsc{explain},
\textsc{query},\textsc{reply}\}.
\]
Only turns mapped to these five moves are used as prediction or generation targets. Excluded turns remain in the preceding dialogue context unless otherwise noted.

\begin{table*}[t]
\centering
\footnotesize
\setlength{\tabcolsep}{3pt}
\renewcommand{\arraystretch}{1.12}
\begin{tabularx}{\textwidth}{
@{}l
>{\raggedright\arraybackslash}X
>{\raggedright\arraybackslash}X
>{\raggedright\arraybackslash}X
>{\raggedright\arraybackslash}X
>{\raggedright\arraybackslash}X
>{\raggedright\arraybackslash}X@{}}
\toprule
\textbf{Dataset}
& \textbf{Acknowledge}
& \textbf{Check}
& \textbf{Explain}
& \textbf{Query}
& \textbf{Reply}
& \textbf{Excluded} \\
\midrule

MapTask
& acknowledge, ready
& check
& explain, clarify
& query\_yn, query\_w
& reply\_y, reply\_n, reply\_w
& align, uncodable \\

FindTask
& acknowledge
& check
& explain
& query\_yn, query\_w
& reply\_y, reply\_n, reply\_w
& align, state, state\_y, state\_n, instruct \\

CReST
& acknowledge, ready, agree, continue
& check
& explain, clarify, assessment
& query\_yn, query\_w
& reply\_y, reply\_n, reply\_w, confirm
& align, instruct, state\_change, incomplete, open, close, 3rd\_turn,
greeting, read, alarm \\

\bottomrule
\end{tabularx}
\caption{Mapping from corpus-specific dialogue-act labels to the shared five-move taxonomy. Excluded labels are not used as target turns but are retained as context unless removed by the preprocessing rules below.}
\label{tab:label-mapping}
\end{table*}

\subsection{Text and Audio Preprocessing}

\paragraph{Turn filtering.}
We remove giver, elder, and director turns from the target set, although they remain available as dialogue context. We additionally exclude turns with empty text and follower turns with no preceding context. MapTask turns labeled \texttt{uncodable} are removed.

For CReST, filler-only turns are removed from both the target set and dialogue context:
\[
\{\texttt{um},\texttt{uh},\texttt{and},\texttt{un},
\texttt{wait},\texttt{whoops},\texttt{and\_um}\}.
\]
We also repair known transcription errors, remove inline annotations enclosed in braces, brackets, or parentheses, and use the primary label preceding ``\texttt{/}'' when multiple tags are present.

\paragraph{Audio processing.}
MapTask and CReST use audio and transcript input, while FindTask is text-only. Dialogue waveforms are loaded from \texttt{<id>.mix.wav}, converted to mono, and resampled to \(16\,\mathrm{kHz}\).

For generation, the audio segment begins \(0.2\,\mathrm{s}\) before the onset of the immediately preceding turn and ends at the onset of the target follower response, ensuring that the held-out response is never included. Segments are capped at \(30\,\mathrm{s}\), retaining the most recent audio when the available context is longer. For teacher-forced activation extraction, the segment extends through the end of the gold follower response.

When a waveform is unavailable, we substitute \(0.1\,\mathrm{s}\) of silence
to preserve the expected number of audio placeholders.

\paragraph{Prompt construction.}
For multimodal conditions, the user input contains the audio item followed by the corresponding transcript. Transcript context contains the most recent \(n\) turns when a context limit is specified and the full preceding dialogue otherwise. Audio+text and text-only conditions are stored and evaluated separately. The automatic move classifier always receives text-only input.

\paragraph{Speaker normalization.}
Corpus-specific roles are rendered in generation prompts using \texttt{G:} for the giver role and \texttt{F:} for the follower role. For automatic classification, these prefixes are further anonymized:
\[
\texttt{G:}\rightarrow\texttt{A:},
\qquad
\texttt{F:}\rightarrow\texttt{B:}.
\]
The speaker prefix is removed from the final response being classified, reducing dataset- and role-specific cues.

\section{Controller Details}
\label{app:controller}

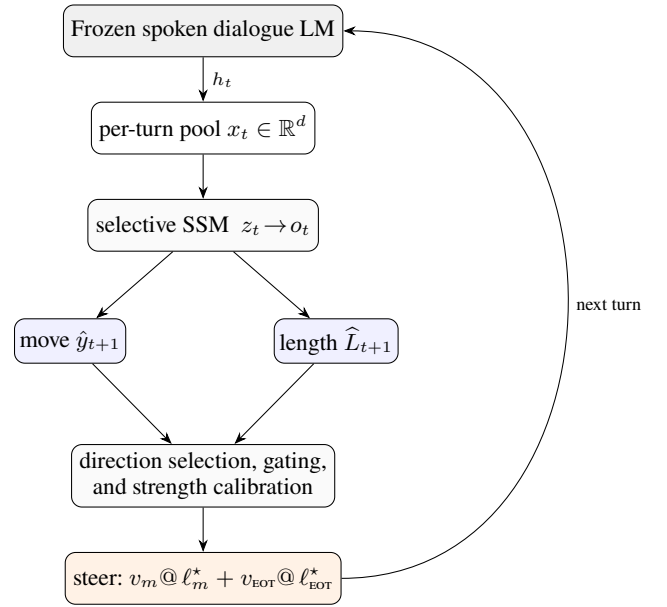
\begin{figure}[h]
\centering
\resizebox{\columnwidth}{!}{%
\begin{tikzpicture}[
  font=\small,
  box/.style={
    draw,
    rounded corners,
    align=center,
    inner sep=3pt,
    minimum height=7mm
  },
  head/.style={
    draw,
    rounded corners,
    align=center,
    inner sep=2pt,
    fill=blue!6,
    minimum height=6mm
  },
  >={Stealth[]},
  node distance=6mm,
]
  \node[box,fill=gray!12] (lm)
    {Frozen spoken dialogue LM};

  \node[box,below=of lm] (feat)
    {per-turn pool $x_t\in\mathbb{R}^d$};

  \node[box,below=of feat,fill=gray!4] (ssm)
    {selective SSM\ \ $z_t\!\to\!o_t$};

  \node[head,below=9mm of ssm,xshift=-18mm] (move)
    {move $\hat y_{t+1}$};

  \node[head,below=9mm of ssm,xshift=18mm] (len)
    {length $\widehat L_{t+1}$};

  \node[box,below=10mm of ssm,yshift=-16mm,fill=gray!4] (policy)
    {direction selection, gating,\\and strength calibration};

  \node[box,below=of policy,fill=orange!10] (steer)
    {steer: $v_m @\,\ell_m^\star$
    $+$ $v_{\textsc{eot}} @\,\ell_{\textsc{eot}}^\star$};

  \draw[->] (lm) -- node[right,font=\scriptsize]{$h_t$} (feat);
  \draw[->] (feat) -- (ssm);
  \draw[->] (ssm) -- (move);
  \draw[->] (ssm) -- (len);
  \draw[->] (move) -- (policy);
  \draw[->] (len) -- (policy);
  \draw[->] (policy) -- (steer);
  \draw[->] (steer.east) to[out=0,in=0,looseness=1.4]
    node[right,font=\scriptsize]{next turn} (lm.east);
\end{tikzpicture}%
}
\caption{
The controller reads the frozen backbone's residual stream and predicts the
next conversational move and response length. The predicted move selects a
move direction, its fixed steering strength, and the realization-aware gate;
the predicted length is calibrated to a turn-boundary coefficient. All
decisions depend on controller predictions rather than gold labels at
inference.
}
\label{fig:controller}
\end{figure}

\paragraph{Per-turn feature.}
For each dialogue turn \(t\), we mean-pool the residual-stream activations at
controller layer \(\ell_{\mathrm{ctrl}}^\star\) over the turn's token span and
standardize them using training-set statistics:
\begin{equation}
x_t
=
\left(
\phi\!\left(h_t^{\ell_{\mathrm{ctrl}}^\star}\right)-\mu
\right)
\oslash \sigma
\in\mathbb{R}^d,
\end{equation}
where \(\phi\) denotes mean pooling, \((\mu,\sigma)\) are the per-coordinate
training-set mean and standard deviation, and \(\oslash\) denotes elementwise
division.

\paragraph{Objective.}
For each retained follower turn \(t\), the causal state after turn \(t-1\)
predicts its move and log response length:
\begin{multline}
\mathcal{L}
=
\sum_{t\in\mathcal{I}_{\mathrm{train}}}
\Big(
\mathrm{CE}\!\left(W_o o_{t-1},y_t\right) \\
+
\lambda\,
\mathrm{smooth}L_1\!\left(
w_L^\top o_{t-1},L_t
\right)
\Big),
\end{multline}
where
\[
L_t
=
\log\!\left(1+\operatorname{words}(u_t)\right),
\]
and \(\mathcal{I}_{\mathrm{train}}\) indexes retained training-set follower
turns whose labels belong to \(\mathcal{M}\).

\paragraph{Length calibration.}
At inference, the predicted log length
\[
\widehat L_{t+1}=w_L^\top o_t
\]
is mapped to a turn-boundary steering coefficient \(\beta\) by a monotone
calibration fitted on the training dialogues. The coefficient is applied only
to naturally terse moves. For content-bearing moves
(\textsc{check} and \textsc{explain}), it is set to zero because shortening
strong enough to affect these moves can overwhelm the move direction.

\paragraph{Streaming inference.}
At deployment, the controller performs one incremental update
\[
(o_t,z_t)
=
\operatorname{step}(x_t,z_{t-1})
\]
per dialogue turn. Its recurrent state persists across turns, so the controller
does not recompute its previous states. The update has constant time and memory
with respect to dialogue history and is equivalent to the parallel scan used
during training, up to a numerical difference below \(10^{-7}\). Thus, the
controller operates as a streaming policy rather than an offline classifier.

\section{Steering Implementation}
\label{app:steering}

\paragraph{Onset features.}
For each retained follower turn \(t\in\mathcal{I}_{\mathrm{train}}\), we extract
the move-onset delta
\[
\Delta_t^\ell
=
h_t^\ell(s_{t,1})
-
h_t^\ell(s_{t,1}-1),
\]
the change in the layer-\(\ell\) residual stream between the positions
immediately before and at the first response token
(Eq.~\ref{eq:move-onset}). Activations are extracted under teacher forcing.
For each target move \(m\), we form a balanced one-vs-rest set containing all
turns labeled \(m\) and an equal-size random sample of the remaining retained
follower turns.

\paragraph{Direction estimation (deployed).}
The deployed steering direction is the PCA-reduced, shrinkage-LDA (Fisher)
direction of Eq.~\ref{eq:lda}, mapped back to the raw residual stream. On the
balanced one-vs-rest set we (i)~standardize the onset features per dimension
(\texttt{StandardScaler}), (ii)~reduce to $k=\min(128,\,n{-}1,\,d)$ principal
components, and (iii)~fit
\texttt{LinearDiscriminantAnalysis(solver="lsqr", shrinkage="auto")}, whose
\texttt{lsqr} coefficient vector is proportional to
$\widehat{\Sigma}_{w}^{-1}(\nu_m^\ell-\nu_{\lnot m}^\ell)$ with the within-class
covariance $\widehat{\Sigma}_{w}$ regularized by automatic Ledoit--Wolf shrinkage
\citep{ledoit2004well}. Let $w^\ell_m$ be this direction in standardized-PCA
coordinates. We map it back to raw activation space, undoing the PCA rotation and
the per-dimension scaling,
\begin{equation}
\tilde v_m^\ell
=
\operatorname{diag}\!\left((\sigma^\ell)^{-1}\right)
(P^\ell)^\top w_m^\ell,
\qquad
v_m^\ell
=
\frac{\tilde v_m^\ell}
{\lVert \tilde v_m^\ell\rVert_2},
\end{equation}
where $P^\ell$ are the PCA components and $\sigma^\ell$ the per-dimension standard
deviations of the standardizer (elementwise division). The result is unit-normed
in the raw residual space ($d\!=\!3584$ for Qwen2.5-Omni). The vector is oriented
toward the target move by construction (LDA class~1 $=$ move), so no separate
sign-fixing rule is applied. Directions are estimated independently per backbone.

\paragraph{Injection.}
Steering is applied additively to the post-block residual stream at the selected
move-specific layer. For every generated token position \(s\),
\begin{equation}
h^{\ell_m^\star}(s)
\leftarrow
h^{\ell_m^\star}(s)
+
\alpha_m^\star
c_{\ell_m^\star}
v_m^{\ell_m^\star}.
\end{equation}
Here, \(\alpha_m^\star\) is the selected move-specific steering coefficient and
\(c_{\ell_m^\star}\) is the median \(L_2\) norm of the residual stream at
layer \(\ell_m^\star\), estimated over generated tokens from \(300\) MapTask
training contexts. This
normalization makes \(\alpha_m^\star\) dimensionless. The intervention is applied
only during decoding, not to prompt-prefill positions, and uses constant strength
across generated tokens with no positional decay.

\paragraph{Layer and strength selection.}
For each move \(m\) and backbone, we select
\(\ell_m^\star\) and \(\alpha_m^\star\) using only MapTask training dialogues
through a two-stage sweep. Stage~1 evaluates all layers at a fixed
\(\alpha=0.1\) and selects the layer maximizing the mean paired
\emph{soft lift}
\[
\mathbb{E}_{c\in\mathcal{C}_{\neg m}^{\mathrm{train}}}
\left[
P_{\mathrm{steered}}(m\mid c)
-
P_{\mathrm{base}}(m\mid c)
\right],
\]
where \(\mathcal{C}_{\neg m}^{\mathrm{train}}\) contains MapTask training
contexts whose gold move is not \(m\), and \(P\) is obtained from the
Qwen2.5-72B move classifier. Stage~2 fixes the selected layer and sweeps
\(\alpha\in\{0.05,0.1,0.2,0.3,0.4\}\), choosing the value with the greatest
soft lift among those whose degenerate-generation rate is at most \(0.10\).
Directions and hyperparameters are estimated independently for each backbone.
The resulting \(v_m^{\ell_m^\star}\), \(\ell_m^\star\), and
\(\alpha_m^\star\) are then transferred unchanged to held-out MapTask,
FindTask, and CReST.

\paragraph{Adaptive strength.}
As an optional variant, we modulate the move-steering strength for each example
using its alignment with the target direction:
\[
\operatorname{scale}
=
\alpha_m^\star c_{\ell_m^\star}
\left(1+\gamma z_{c,m}\right),
\]
where \(z_{c,m}\) is the alignment of the context representation at
\(\ell_m^\star\) with \(v_m^{\ell_m^\star}\), standardized using the training-set
mean and standard deviation. The coefficient \(\gamma\) controls the adaptive
gain; \(\gamma=0\) recovers the fixed-strength intervention.

\section{Turn-Boundary Subspace and Realization-Aware Gating}
\label{app:eot}

\paragraph{Predicted length.}
The controller regresses the follower turn's log length
\[
L_{t+1}
=
\log\!\left(1+\operatorname{words}(u_{t+1})\right).
\]
At inference, the prediction
\[
\widehat L_{t+1}=w_L^\top o_t
\]
is mapped to the turn-boundary coefficient \(\beta\) by a monotone calibration
fitted on training dialogues.

\begin{table}[t]
\centering
\small
\begin{tabular}{lcc}
\toprule
\textbf{Predicted Move}
& \textbf{Move Steering}
& \textbf{EOT Steering} \\
\midrule
acknowledge & apply    & apply \\
check       & apply    & withhold \\
explain     & apply    & withhold \\
query       & apply    & apply \\
reply       & withhold & apply \\
\bottomrule
\end{tabular}
\caption{Realization-aware gating policy.}
\label{tab:gating}
\end{table}

\section{Human Study}
\label{app:human_study}

We conduct a human study to validate the automatic move classifier against human judgment on model-generated responses, and to measure agreement between human annotators and the LLM-based evaluator. From the pooled generations of all methods, we sample \(500\) responses---\(100\) for each of the five target moves, distributed approximately equally across the three generation models and three datasets. Each response is annotated by three workers who see only the dialogue context and the generated utterance and select one conversational move; the target move, generation method, and classifier prediction are withheld so that annotation is blind.

\paragraph{Annotators.}
All annotations were collected on Amazon Mechanical Turk (AMT) using workers with a minimum approval rate of 95\%. Annotators were restricted to English-speaking countries to ensure reliable comprehension. Compensation was set to meet or exceed the U.S.\ federal minimum wage based on conservative task-time estimates. No personally identifiable information was collected.

\subsection{Human--LLM Agreement}

We compare human judgments to the LLM move classifier at the item level. Human decisions are aggregated by plurality voting across the three annotators, and the LLM prediction is the label assigned by the two-stage Qwen2.5-72B-Instruct classifier \citep{yang2025qwen3}. Agreement is reported per dataset in Table~\ref{tab:human_llm_agreement}: overall, the classifier matches the human plurality on \(73.9\%\) of items, rising to \(90.5\%\) on the \(327\) items where all three annotators agreed. Inter-annotator agreement is moderate (Fleiss' \(\kappa = 0.682\) overall), and the classifier's agreement with the human majority is comparable to the agreement among annotators themselves.

\begin{table}[h]
\centering
\footnotesize
\setlength{\tabcolsep}{3.5pt}
\renewcommand{\arraystretch}{1.08}
\begin{tabular}{lcc}
\toprule
\textbf{Dataset}
& \shortstack{\textbf{Plurality}\\\textbf{Agreement}}
& \shortstack{\textbf{Unanimous}\\\textbf{Agreement}} \\
\midrule
MapTask  & 76.4\% & 94.6\% \\
FindTask & 67.8\% & 91.9\% \\
CReST    & 75.5\% & 86.5\% \\
\midrule
Overall  & 73.9\% & 90.5\% \\
\bottomrule
\end{tabular}
\caption{Agreement between the automatic classifier and human judgments. Plurality agreement uses items with a clear majority; unanimous agreement uses only items on which all three annotators selected the same move.}
\label{tab:human_llm_agreement}
\end{table}

To assess alignment at the distribution level, we compare the classifier's normalized move probabilities with the human vote shares across all item--move pairs. The two are positively correlated (Pearson \(r = 0.721\) overall, ranging from \(r = 0.62\) for \textit{explain} to \(r = 0.85\) for \textit{query}), indicating that the classifier's confidence tracks human uncertainty. Moreover, ranking the seven generation methods by human-judged target-move realization reproduces the classifier-based ranking almost exactly (Spearman \(\rho = 0.96\), Pearson \(r = 0.99\)), showing that the two evaluators agree not only per item but also in their relative assessment of methods. These results support the use of the LLM move classifier as a scalable proxy for human judgment in our evaluation.

\subsection{MTurk Interfaces}

\begin{figure*}[h]
\centering
\includegraphics[width=0.8\linewidth]{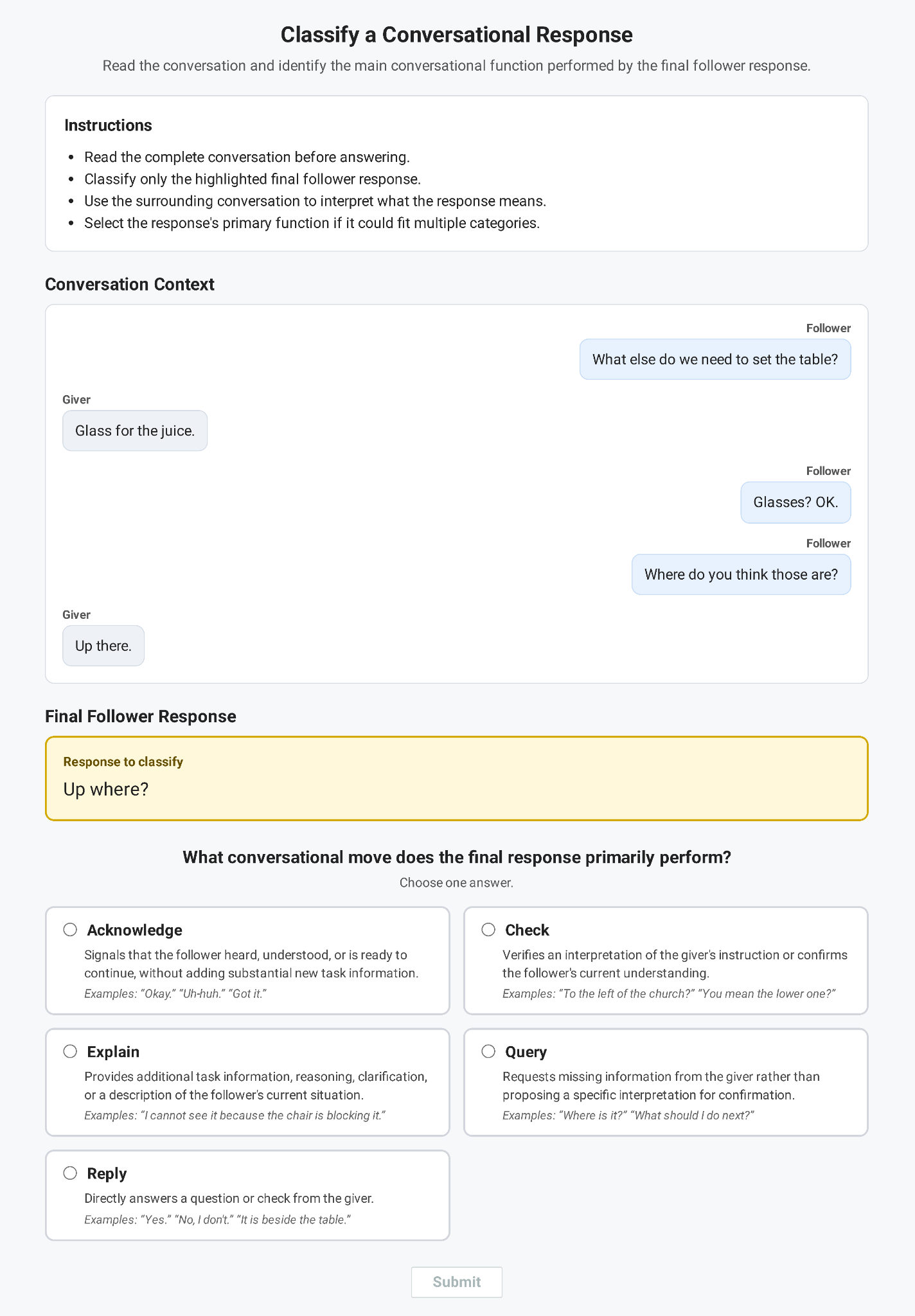}
\caption{
Human annotation interface for move classification. Annotators are shown the preceding dialogue context and a single candidate response, and asked to select the conversational move the response performs (acknowledge, check, explain, query, or reply). The target move, generation method, and classifier prediction are hidden.
}
\label{fig:human_eval_ui}
\end{figure*}

Figure~\ref{fig:human_eval_ui} shows the annotation interface presented to workers. 

